\documentclass[runningheads]{llncs}
\usepackage[T1]{fontenc}
\usepackage{multirow}
\usepackage{graphicx}
\usepackage{booktabs}
\usepackage{float}
\usepackage{hyperref}
\usepackage{diagbox}
\usepackage{subcaption}
\usepackage[table]{xcolor}
\usepackage[font=small,labelfont=bf]{caption}

\begin{document}
\title{Advancing SEM Based Nano-Scale Defect\\Analysis in Semiconductor Manufacturing\\for Advanced IC Nodes}
\titlerunning{Advancing SEM Defect Analysis for Advanced IC Nodes}

\author{Bappaditya Dey\inst{1,*} \and
Matthias Monden\inst{1,2,*} \and
Victor Blanco\inst{1} \and
Sandip Halder\inst{1} \and
Stefan De Gendt\inst{1,2}
}
\authorrunning{B. Dey et al.}

\institute{imec, Kapeldreef 75, 3000 Leuven, Belgium\inst{1} \\
KU Leuven, Oude Markt 13, 3000 Leuven, Belgium\inst{2}\\
Equal Contribution\inst{*} \\
\email{Bappaditya.Dey@imec.be}}
\maketitle              
\begin{abstract}
    In this research, we introduce a unified end-to-end Automated Defect Classification-Detection-Segmentation (ADCDS) framework for classifying, detecting, and segmenting multiple instances of semiconductor defects for advanced nodes. 
This framework consists of two modules: (a) a defect detection module, followed by (b) a defect segmentation module.
The defect detection module employs Deformable DETR to aid in the classification and detection of nano-scale defects, while the segmentation module utilizes BoxSnake. 
BoxSnake facilitates box-supervised instance segmentation of nano-scale defects, supported by the former module. 
This simplifies the process by eliminating the laborious requirement for ground-truth pixel-wise mask annotation by human experts, which is typically associated with training conventional segmentation models. 
We have evaluated the performance of our ADCDS framework using two distinct process datasets from real wafers, as ADI and AEI, specifically focusing on Line-space patterns.
We have demonstrated the applicability and significance of our proposed methodology, particularly in the nano-scale segmentation and generation of binary defect masks, using the challenging ADI SEM dataset where ground-truth pixel-wise segmentation annotations were unavailable.
Furthermore, we have presented a comparative analysis of our proposed framework against previous approaches to demonstrate its effectiveness. 
Our proposed framework achieved an overall mAP@IoU0.5 of 72.19 for detection and 78.86 for segmentation on the ADI dataset. Similarly, for the AEI dataset, these metrics were 90.38 for detection and 95.48 for segmentation. 
Thus, our proposed framework effectively fulfils the requirements of advanced defect analysis while addressing significant constraints.
\end{abstract}
\section{Introduction}
The scaling of semiconductor circuits has yielded various advantages, including enhanced device speed and reduced power consumption. This trend of scaling (towards sub-30nm pitches for 5nm node and below), spanning the past fifty years, is aptly represented by Moore's Law. 
However, to maintain this trajectory of scaling, it is imperative to introduce novel concepts for manufacturing processes. 
Currently, industry is evaluating the use of High-Numerical Aperture Extreme Ultraviolet Lithography (High-NA EUVL) \cite{high_na_challenges} technology for reducing further the pitch in future nodes. 
One of the main challenges in introducing High-NA EUV in High Volume Manufacturing (HVM) is its low depth of focus which is pushing resist material suppliers to use thinner resists \cite{10.1117/12.2643315} and new underlayers/hardmask's. 
When thin resist materials are used in conjunction with novel underlayers and hardmask's, often it poses signal detection challenges for inspection and metrology equipment's due to low Signal-to-Noise Ratio (SNR).
In such a scenario, detection of these tiny defects becomes extremely challenging. 
Furthermore, manual classification and detection of these nano-scale defects accurately and consistently in the acquired Scanning Electron Microscope (SEM) images is nearly impossible for humans. 
Moreover, industrial defect-inspection tools face challenges in consistently enhancing sensitivity and scalability to address the aforementioned issues. 
Neither approach is sustainable in the long run. Under these circumstances, deep learning based methods \cite{retinanet} are becoming more and more useful to enhance signal from the tiny defects/perturbations. 
Furthermore, as increasingly powerful CPU/GPU units emerge and various deep learning (DL) models continue to develop, their implementation is becoming increasingly ubiquitous. 
When implemented correctly, these models enhance defect detection efficiency, maintain scalability, and achieve heightened accuracy/precision, thus reducing the necessity for human intervention.
Supervised algorithms necessitate meticulous manual data labelling, adding extra time and effort, especially when annotating accurate pixel-wise segmentation masks for training models focused on nano-scale defect instance segmentation. 
Alternatively, unsupervised strategies may offer significant advantages, particularly if we can avoid the need for manual bounding-box labelling or precise pixel-wise segmentation mask labelling during model training. 
This approach could lead to more efficient and scalable solutions for defect analysis in semiconductor manufacturing processes, facilitating unsupervised defect classification, detection, and segmentation. \par
The aim of this research is to advance towards unsupervised defect localization and segmentation by introducing a novel ADCDS framework. 
The defect dataset utilized in this study is obtained from real FAB processes, specifically post-litho (After-Development-Inspection) and post-etch (After-Etch-Inspection) stages. 
Examples for different defects are given in figure \ref{fig:adi_defects} and figure \ref{fig:aei_defects} for these two SEM datasets. Unlike some previous studies that rely on digital twins or synthetic datasets, we opted to focus on real FAB data to address the challenges posed by stochastic defectivity scenarios. 
Additionally, we avoided using fabricated datasets with intentionally placed or programmed defect types. Our research contributions are as follows:
\begin{itemize}
    \setlength{\itemsep}{0pt}
    \item Introduction of a unified end-to-end framework for classifying, detecting, and segmenting semiconductor defect instances in aggressive pitches.
    \item Adoption of the Deformable DETR architecture, a query-based framework employing deformable convolution to increase convergence speed and sensitivity to smaller defect features, as the primary defect detection module for classification and detection of nano-scale defects. This information is then seeded into the segmentation module for further processing.
    \item Integration of the BoxSnake architecture in the segmentation module to eliminate the requirement for laborious ground-truth pixel-wise mask annotation by human experts. This alleviates the difficulty associated with annotating pixel-level stochastic defects accurately enough to train a supervised model for deployment in high-volume manufacturing (HVM).
    \item Generation of pixel-precise segmented binary masks for various inter-class and intra-class defect instances, facilitating advanced data analytics to aid improved semiconductor process control.
    \item Benchmarking of the proposed ADCDS framework on two process datasets, namely ADI- and AEI- SEM datasets. Additionally, a first demonstration on the challenging ADI SEM dataset, where ground-truth pixel-wise segmentation annotations were unavailable, is presented.
\end{itemize}
\section{Related Work} \label{sec:relatedwork}
This section aims to briefly discuss selected previous research works that will serve as a baseline for our proposed framework in this study. Our selection criteria are based on: 
(1) previous studies utilizing real resist wafer datasets from FAB for benchmarking on different process steps, primarily After-Development-Inspection (ADI) and After-Etch-Inspection (AEI); (2) studies targeting both defect instance detection and segmentation; and finally, (3) some baseline advanced architectures/strategies that have inspired our proposed framework.\par
Ref. \cite{comparative_study} offers a comparative analysis of state-of-the-art DL-based object detector models for semiconductor defect detection in ADI SEM images. 
This study revisits previously reported applications of RetinaNet \cite{retinanet}, and YOLOv7 \cite{yolov7} models on semiconductor defect detection. Furthermore, the authors introduce several other benchmarks utilizing models such as YOLOv7x, an architecture variant of YOLOv7 \cite{yolov7} with larger dimensions aimed at preserving speed, DINO \cite{dino_org}, and Faster R-CNN \cite{fasterrcnn}.
In Ref. \cite{yolov7_adi}, the authors optimized the YOLOv7 model architecture for semiconductor defect detection by manually adjusting hyperparameters and merging predictions from various models using the WBF (Weighted Box Fusion\footnote{\url{https://github.com/ZFTurbo/Weighted-Boxes-Fusion}}) method, replacing the conventional NMS method (Non-Maximum Suppression).
Only one framework places emphasis on defect detection in AEI SEM images, namely SEMI-Centernet \cite{semi_centernet}, which primarily focuses on inference speed. \par
To assess the performance of the semiconductor defect instance segmentation module, examined two baseline frameworks are: SEMI-PointRend \cite{semi_pointrend}, and SEMI-DiffusionInst \cite{semi_diffusioninst}. 
Although the Mask R-CNN approach \cite{mask_rcnn_approach} provides valuable insights that are considered in this study, it is the initial model to have been applied to the AEI SEM dataset and therefore will not be included in our comparison.
PointRend \cite{pointrend} tries to solve image segmentation as classical rendering problem in computer graphics, resulting in far more efficient computations and allowing higher resolution input images. 
Hence, SEMI-PointRend is capable of creating more clear boundaries for the segmentation masks resulting in higher precision on the masks compared to the standard Mask R-CNN framework \cite{mask_rcnn_approach}.
DiffusionInst \cite{diffusioninst} treats image segmentation as a gradual denoising process, starting from a constructed noisy image. SEMI-DiffusionInst is the first diffusion-like model applied on SEM images for defect segmentation \cite{semi_diffusioninst}. 
Reportedly, SEMI-DiffusionInst \cite{semi_diffusioninst} surpasses the previous leading SEM defect segmentation model, SEMI-Pointrend \cite{semi_pointrend}, when benchmarked on the same AEI SEM dataset. \\ 
The references mentioned above exclusively feature supervised segmentation models, necessitating a ground-truth segmentation mask (manually labeled pixel-wise annotation). These ground-truth masks for the AEI SEM dataset were initially introduced in \cite{mask_rcnn_approach}.
To date, no models have been employed to segment ADI SEM images; hence, no discussions can be made on this dataset. \par
The concept of utilizing detection transformers, particularly queries, for instance segmentation is not new. For instance, `Segment Objects by Learning Queries' (SOLQ) \cite{solq} leverages Deformable DETR and a new Unified Query Representation (UQR) module to enable end-to-end object detection and segmentation.
Similarly, in the paper `Instances as Queries' \cite{queryinst}, authors the authors address the challenge of integrating DETR (and other variants) into Mask R-CNN by introducing QueryInst. QueryInst presents a novel multi-stage instance segmentation network that incorporates a new dynamic convolution mask head architecture on top of a query based end-to-end object detector for image segmentation.
The authors effectively demonstrate parallel supervision on the mask heads by using queries greatly improve instance segmentation compared to non-query based frameworks. 
Masked-attention Mask Transformer (Mask2Former) \cite{mask2former} attempts to establish a universal query-based segmentation architecture that outperforms specialised models like QueryInst or SOLQ. \par
In this work we employ two modules in the proposed ADCDS framework: (1) the Deformable DETR \cite{deformable_detr}, and (2) BoxSnake \cite{boxsnake}. 
The Deformable DETR greatly improves convergence speed and small object detection over the original DETR \cite{detr} by introducing the deformable attention module. 
BoxSnake is the first box-supervised DL-framework built on top of Mask R-CNN, capable of segmenting objects.
\vspace{-0.4cm}
\section{Proposed ADCDS framework} \label{sec:methodology}
\vspace{-0.2cm}
In this section, we briefly outline our proposed unified framework for detecting and segmenting nano-scale defect instances in both ADI- and AEI- SEM images. Our approach relies exclusively on bounding-box ground-truth annotations, aiming to achieve unsupervised precise defect instance segmentation and mask generation.
\begin{figure}[htp]
    \centering
    \includegraphics[width=7.5cm]{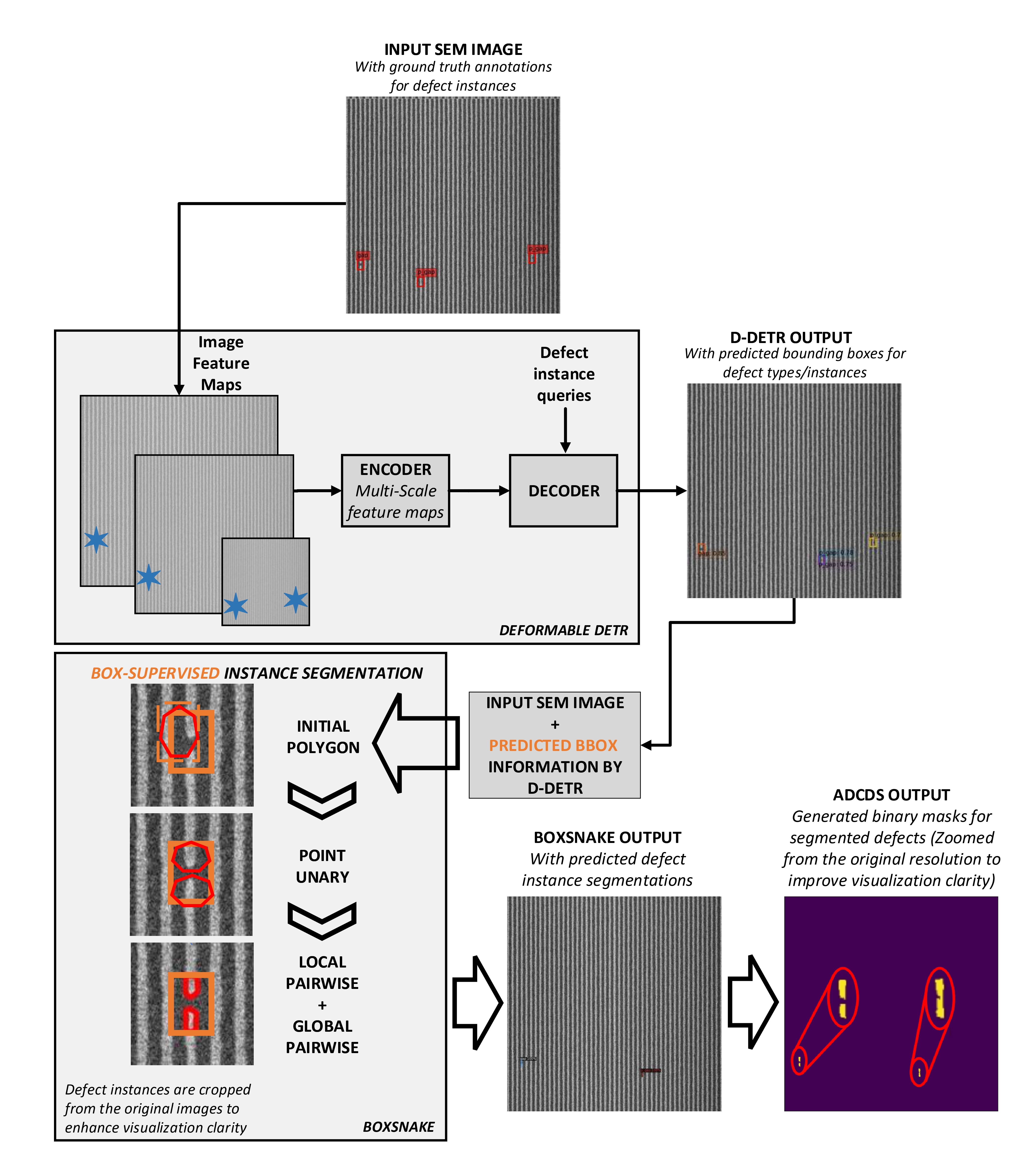}
    \caption{Schematic of the proposed ADCDS framework}
    \label{fig:adcds_framework}
\end{figure}
\newline The schematic of our proposed ADCDS framework pipeline is depicted in figure \ref{fig:adcds_framework}, comprising two modules: (a) a defect detection module, followed by (b) a defect segmentation module. 
The defect detection module utilizes the Deformable DETR \cite{deformable_detr} architecture primarily for classifying and detecting nano-scale defects. 
At this stage, the module takes SEM images with corresponding ground-truth bounding box annotations (manually annotated by a defect inspection expert) for training. 
Once we validate the model's robustness and generalizability on the same defect patterns/types, we deploy it to generate defect classification, detection, and localization information (in terms of bounding box coordinates) for new SEM image batches from similar processes (both ADI and AEI), automating the defect detection mechanism. 
Subsequently, we feed this defect classification and detection result (bounding box information) as the Region-of-Interest (ROI) to the segmentation module, which utilizes the BoxSnake \cite{boxsnake} architecture. 
This architecture enables box-supervised instance segmentation of nano-scale defects, simplifying the process by eliminating the laborious requirement for ground-truth pixel-wise mask annotation by human experts, 
typically associated with training conventional segmentation models. After training, the segmentation module predicts precise segmentation masks for each defect instance. 
Therefore, our proposed ADCDS framework assists in accurately segmenting the extent of defect patterns and aids in distinguishing between various types of inter-class and intra-class stochastic defect patterns 
(such as differentiating between a line collapse and a bridge or distinguishing between closely related patterns like multi-bridge horizontal and multi-bridge non-horizontal). 
This segmentation is crucial for conducting root-cause analysis of defect generation, including factors such as process drift and tool deviations. 
The training strategy of the proposed framework, along with each corresponding module, is discussed in subsequent subsection\ref{sec:training}
\section{Experiments}
\subsection{Datasets}
\begin{figure}[h]
    \begin{subfigure}[b]{\textwidth}
        \centering
        \includegraphics[width=0.65\textwidth]{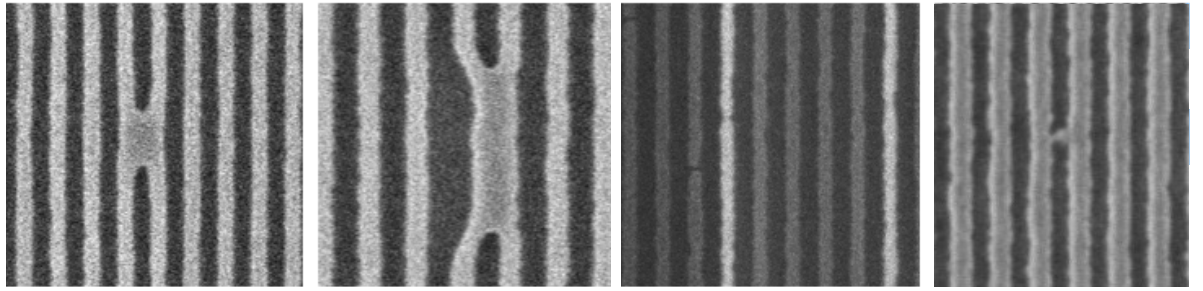}
        \caption{Examples defect classes for the ADI dataset (Line-Space pattern). From left to right: bridge, line collapse, gap and probable gap, and micro bridge}
        \label{fig:adi_defects}
    \end{subfigure}
    \\
    \begin{subfigure}[b]{\textwidth}
        \centering
        \includegraphics[width=\textwidth]{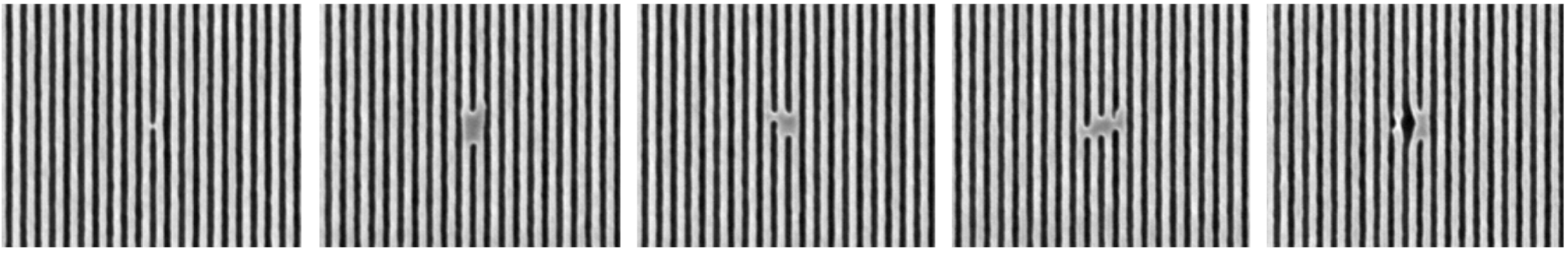}
        \caption{Examples defect classes for the AEI dataset (Line-Space pattern). From left to right: thin bridge, single bridge, multi bridge horizontal, multi bridge non-horizontal, and line collapse}
        \label{fig:aei_defects}
    \end{subfigure}
    \caption{Defect examples from ADI (a) and AEI (b) dataset}
\end{figure}

\vspace{-0.2 cm}
Both ADI and AEI SEM image datasets (for Line-Space pattern) with stochastic defect patterns have been obtained from actual FAB environments.
Raw SEM images are acquired from the imaging tool in ".tiff" format and subsequently converted to ".jpg" format to align with the expected format by the models used. All ADI SEM images have a resolution of 1024x1024, while all AEI SEM images have a resolution of 480x480.
Typical defects found in the ADI dataset include gaps, probable gaps (pgaps), line collapses, bridges, and microbridges, as depicted in figure \ref{fig:adi_defects}.
Figure \ref{fig:aei_defects} illustrates typical defects for the AEI dataset, which include thin bridges, single bridges, multi bridges (non) horizontal (MBH/MBNH), and line collapses.
The distribution of defects for total images and representative defect classes is presented in table \ref{tab:dist_adi} for the ADI dataset and table \ref{tab:dist_aei} for the AEI dataset, respectively.

\begin{table}[b]
    \parbox{0.48\linewidth}
    {
        \caption{Data distribution and defect class information of ADI SEM images}
        \centering
        \fontsize{7.6pt}{9pt}
        \selectfont
        \begin{tabular}{|r || c | c |}
            \hline
            & \textbf{Train} & \textbf{Validation}\\ \hline
            \textbf{Total images} & 1054 & 117 \\ \hline
            \multicolumn{1}{|c||}{\textbf{Classes}} & \multicolumn{2}{c|}{Total instances} \\ \hline
            Gap & 1046 & 156\\
            Probable gap & 315 & 49\\
            Bridge & 238 & 19\\
            Micro bridge & 380 & 47\\
            Line collapse & 550 & 66\\ \hline
            \textbf{Total instances} & 2529 & 337\\
            \hline
        \end{tabular}
        \label{tab:dist_adi}
    }
    \hfill
    \parbox{0.48\linewidth}
    {
        \caption{Data distribution and defect class information of AEI SEM images}
        \centering
        \fontsize{7.5pt}{9pt}
        \selectfont
        \begin{tabular}{|r || c | c |}
            \hline
            & \textbf{Train} & \textbf{Validation}\\ \hline
            \textbf{Total images} & 1062 & 131\\ \hline
            \multicolumn{1}{|c||}{\textbf{Classes}} & \multicolumn{2}{c|}{Total instances} \\ \hline
            Multi bridge NH & 179 & 21\\
            Multi bridge H & 90 & 10\\
            Single bridge & 271 & 29\\
            Thin bridge & 270 & 29\\
            Line collapse & 236 & 40\\ \hline
            \textbf{Total instances} & 1046 & 129\\
            \hline
        \end{tabular}
        \label{tab:dist_aei}
    }
\end{table}

\vspace{-0.5cm}
\subsection{Evaluation criteria}
In the subsequent subsections, the methodology for training and inference results is outlined.
Based on these specifications, results are presented in section \ref{sec:results} and compared with previous related research works \cite{comparative_study,semi_pointrend,semi_diffusioninst,semi_centernet}, in section \ref{sec:discussion}, alongside insights from our proposed approach. 
Key performance metrics of interest include AP (Average Precision), mAP (mean AP), and inference speed.
The AP is computed using the COCO evaluation framework integrated into the Deformable DETR and BoxSnake frameworks. Typically, the AP of the detector/segmentation model is determined using Intersection-over-Union (IoU). 
IoU calculates the intersection of the ground truth annotated bounding box containing the defect instance with the predicted bounding box, divided by the union of both bounding boxes.
The inference speed for the Deformable DETR is determined by the time it takes for an image to propagate through the model, without considering the extraction of results due to hardware and implementation dependencies. 
Inference speed for BoxSnake is taken from the Detectron2 output.
\vspace{-0.5 cm}

\subsection{Training} \label{sec:training}
To ensure comparability and consistency in results, all trainings/simulations and inference runs are performed on an NVIDIA A100 GPU.
\vspace{-0.5cm}
\subsubsection{Defect instance detection. } 
All configurations of Deformable DETR \cite{deformable_detr} utilize a ResNet-50 backbone with COCO pretrained weights. While maintaining hyperparameters as closely as possible to the baseline Deformable DETR implementation, adjustments are made to the learning rate, learning rate drop, and the number of queries of the transformer.
The learning rate for both ADI- and AEI datasets is reduced to $5e-6$ in order to reach AP convergence in a reasonable time window. The learning rate drop is increased from $40$ to $175$.
Lastly, the batch size is set to $2$ and training is done over a period of $500$ epochs. The epoch with the best AP is chosen as benchmark.
\vspace{-0.5cm}
\setlength{\parskip}{0pt}
\subsubsection{Defect instance segmentation. }
In our proposed framework, we explored each baseline configuration outlined in the original BoxSnake \cite{boxsnake} GitHub\footnote{\url{https://github.com/Yangr116/BoxSnake}} repository, providing an overview in Table \ref{tab:boxsnake_configs_overview}. 
We highlighted only the most effective configuration for our research goal. We conducted a total of 10,000 training iterations, as reported by Detectron2. \par
For the ADI dataset, we had ground-truth bounding box annotations but no ground-truth pixel-wise defect mask annotations. Therefore, `segmentation AP' is calculated manually by using the typical precision formula $\frac{\#True Positives}{\#TP + \#False Positives}$, 
where a True Positive segmentation indicates that the model accurately segmenting the defect instance(s) based solely on bbox supervision, while a False Positive indicates instances where the model incorrectly segments defect instances (say, a probable gap as a gap) or segments other parts in the SEM images not corresponding to intended defect instances guided by bbox supervision.
Additionally, the recall metric is based on the following formula: $\frac{\#True Positives}{\#TP + \#False Negatives}$. In this case False Negative indicates the model failed to detect true defect instances.
Only detections with a confidence threshold of 0.7 or higher are considered. ADI SEM images are too noisy to generate precise masks for all instances and require refinement. 
Improved binary masks are generated by creating one from the original image and then performing an AND-operation with the predicted segmented mask result from the BoxSnake module for each defect instance. 
Further refinement is accomplished by coloring a pixel based on the color of its neighboring pixel.
For the AEI dataset, we present (m)AP detection and segmentation results using the same metric calculation strategy as outlined in \cite{mask_rcnn_approach}, as this dataset includes both ground-truth bounding box annotations and pixel-wise defect mask annotations.
\vspace{-0.5cm}

\begin{table}[htp]
    \caption{Benchmark configurations for the segmentation model}
    \centering
    \fontsize{7.5pt}{9pt}
    \selectfont
    \begin{tabular}{|c|c|c|}
        \hline
        \textbf{Dataset}    & \textbf{Backbone}  & \textbf{Learning rate scheduler}\\
        \hline
        \multirow{3}{*}{ADI \& AEI} & ResNet-50 & 1 \& 2\\
        & ResNet-101 & 1 \& 2\\
        & Swin Base \& Large & 1\\
        \hline
    \end{tabular}
    \label{tab:boxsnake_configs_overview}
\end{table}
\vspace{-0.5cm}
\section{Results} \label{sec:results}
\subsection{Proposed ADCDS framework}
In this section, we will discuss and demonstrate the performance of our proposed ADCDS framework, 
along with each corresponding module, in the task of nano-scale defect inspection and segmentation, supported by proof-of-validation.
\vspace{-0.5cm}
\subsubsection{Defect detection module. }
\begin{figure}[htp]
    \begin{subfigure}{\textwidth}
        \centering
        \includegraphics[width=\textwidth]{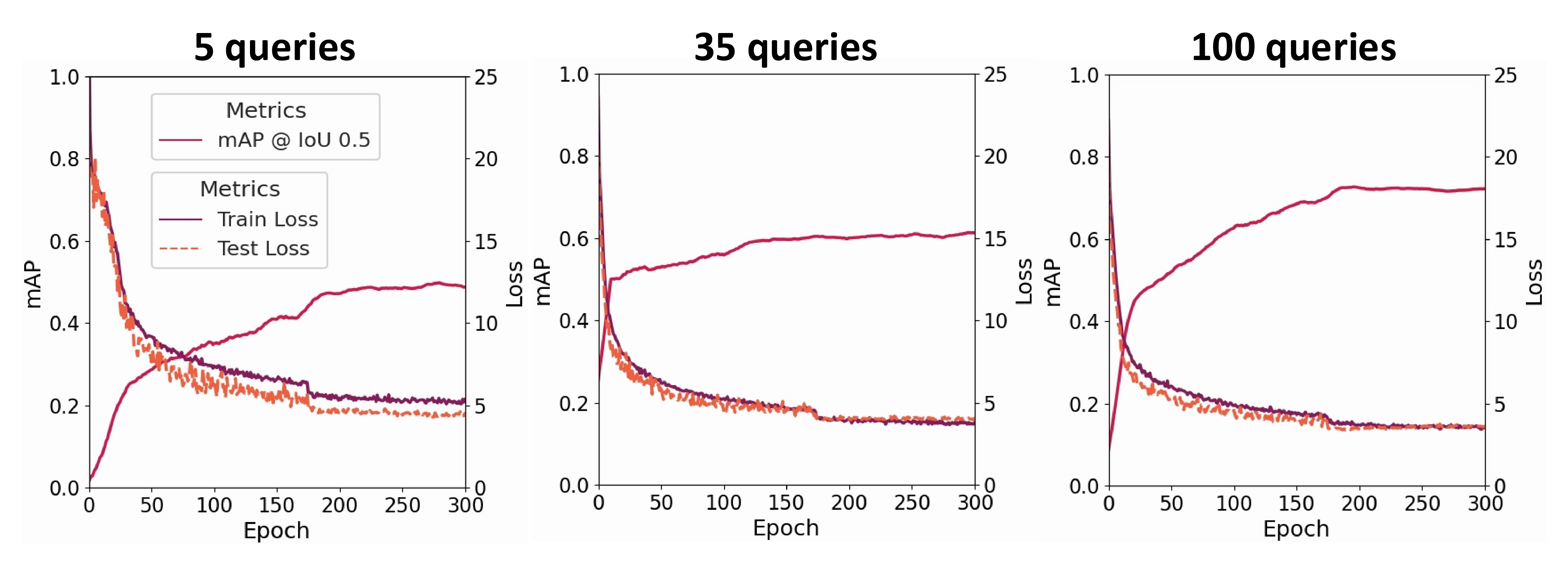}
        \caption{ADI dataset}
        \label{fig:convergence_adi_ddetr}
    \end{subfigure}
    \begin{subfigure}{\textwidth}
        \centering
        \includegraphics[width=0.7\textwidth]{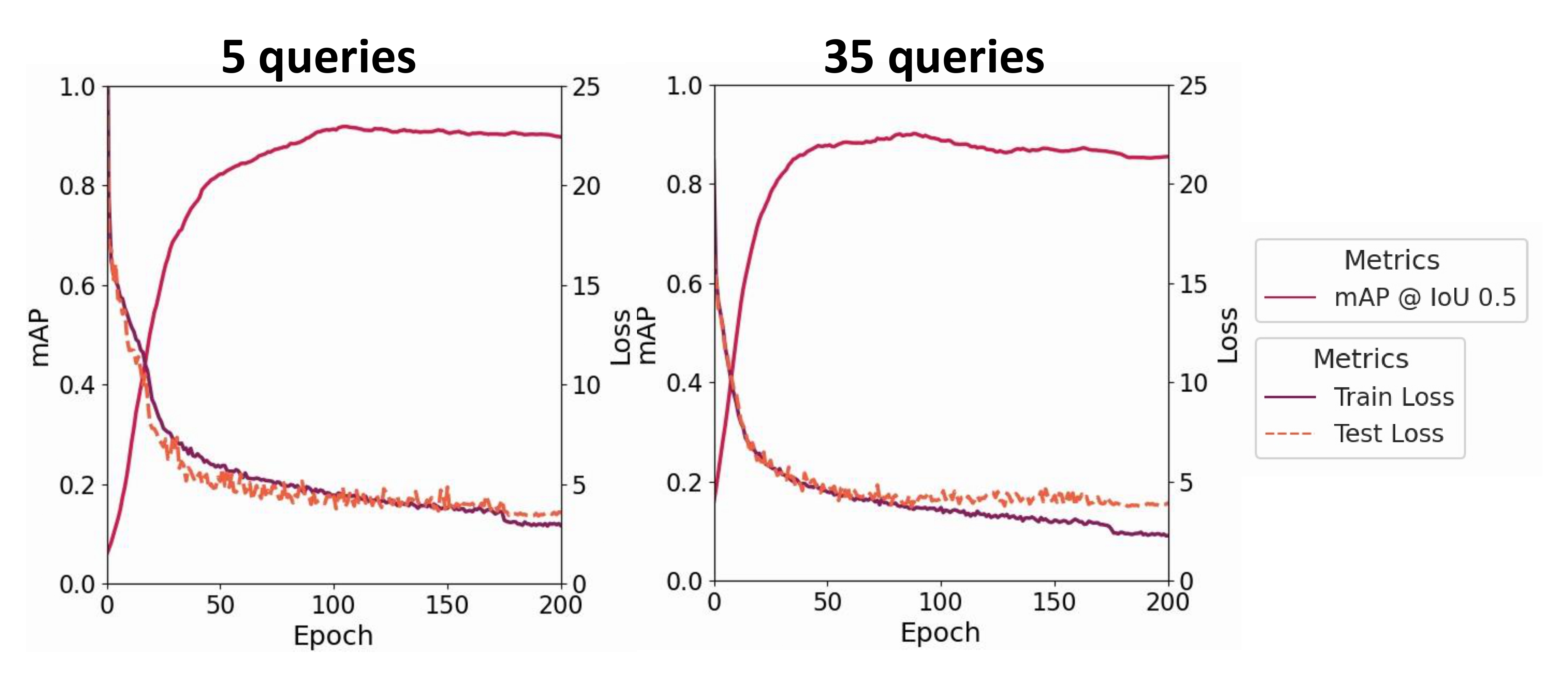}
        \caption{AEI dataset}
        \label{fig:convergence_aei_ddetr}
    \end{subfigure}
    \caption{mAP @ IoU 0.5, train, and test loss with Deformable DETR architectures on the (a) ADI and (b) AEI SEM datasets}
    \label{fig:convergence_ddetr}
\end{figure}

\begin{figure}[htp]
    \centering
    \begin{subfigure}{\textwidth}
        \centering
        \includegraphics[width=0.8\textwidth]{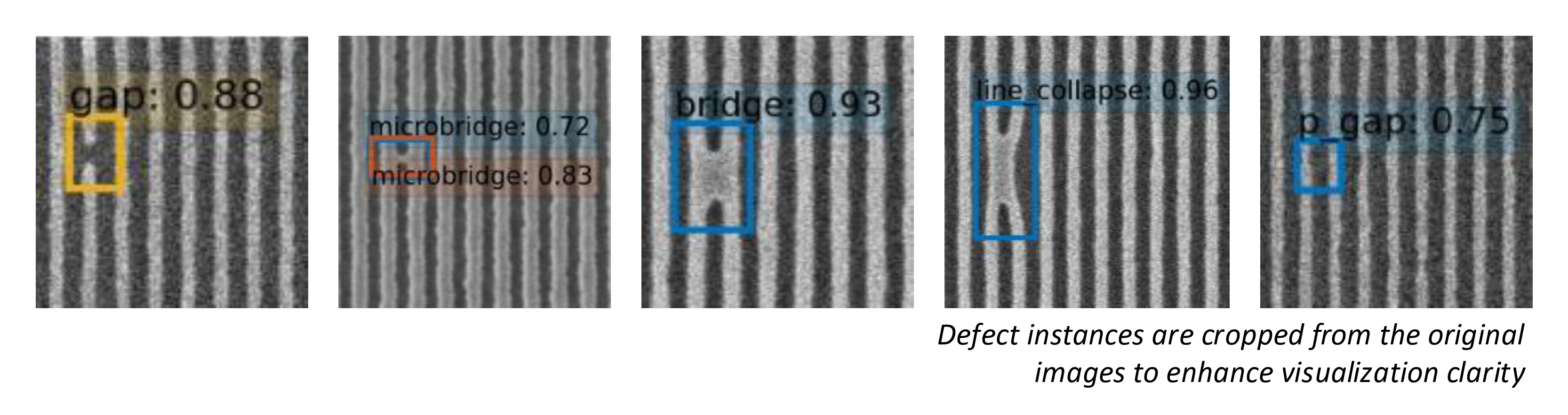}
        \caption{Prediction results on ADI SEM Test dataset}
    \end{subfigure}
    \\
    \begin{subfigure}{\textwidth}
        \centering
        \includegraphics[width=0.8\textwidth]{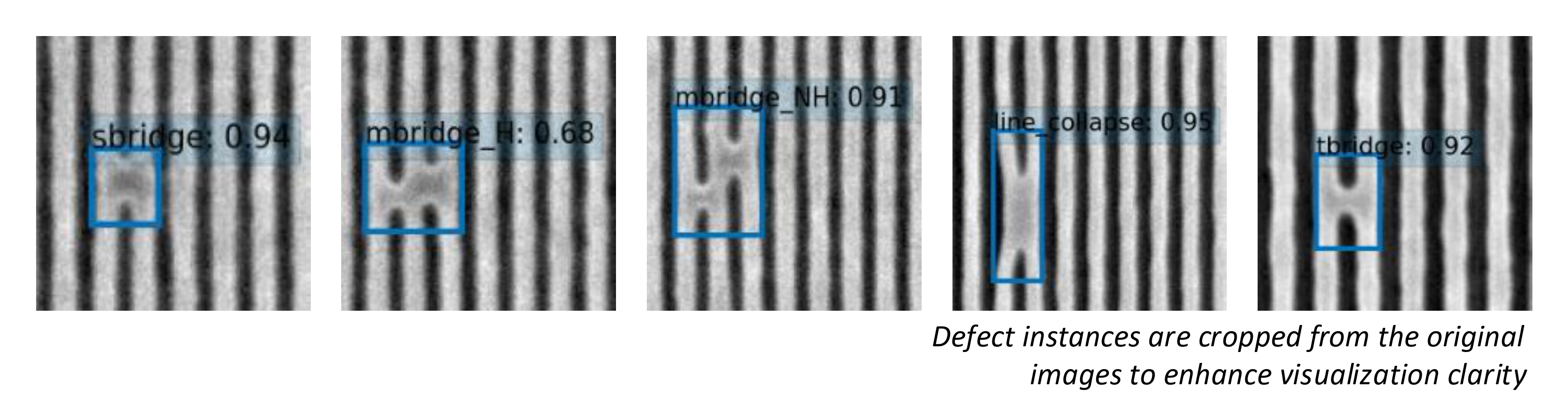}
        \caption{Prediction results on AEI SEM Test dataset}
    \end{subfigure}
    \caption{Defect detection results with proposed ADCDS framework (Deformable DETR based defect detection module) on ADI (a) and AEI (b) SEM dataset}
    \label{fig:defect_detection_visually_adiaei}
\end{figure}

The metrics presented in this section encompass both the ADI and AEI datasets. The per-class AP and mAP for the ADI dataset are presented in table \ref{tab:results_adi} for the detection module, with different query configurations of 5, 35, and 100. All configurations seem to converge after 200 epochs, as shown in figure \ref{fig:convergence_adi_ddetr}.
Likewise, per-class AP and mAP for the AEI dataset are presented in table \ref{tab:results_aei}, with results obtained from two different training epochs (100 and 200) and two query configurations (5 and 35), as shown in figure \ref{fig:convergence_aei_ddetr}. We discarded the query configuration for 100 as it adversely affected the precision metrics.
Among all experimental configurations, we selected the model with the highest mAP score @IoU 0.5 for comparative analysis with previous benchmarks in section \ref{sec:discussion}, for both ADI and AEI datasets.
Table \ref{tab:inferencespeed} presents detection inference timings (ms/image). Examples illustrating defect classification and detection results on both ADI and AEI SEM test datasets are presented in figure \ref{fig:defect_detection_visually_adiaei}.

\vspace{-0.5cm}
\subsubsection{Defect segmentation module. }
\begin{figure}[ht]
    \centering
    \begin{subfigure}{\textwidth}
        \centering
        \includegraphics[scale=0.15]{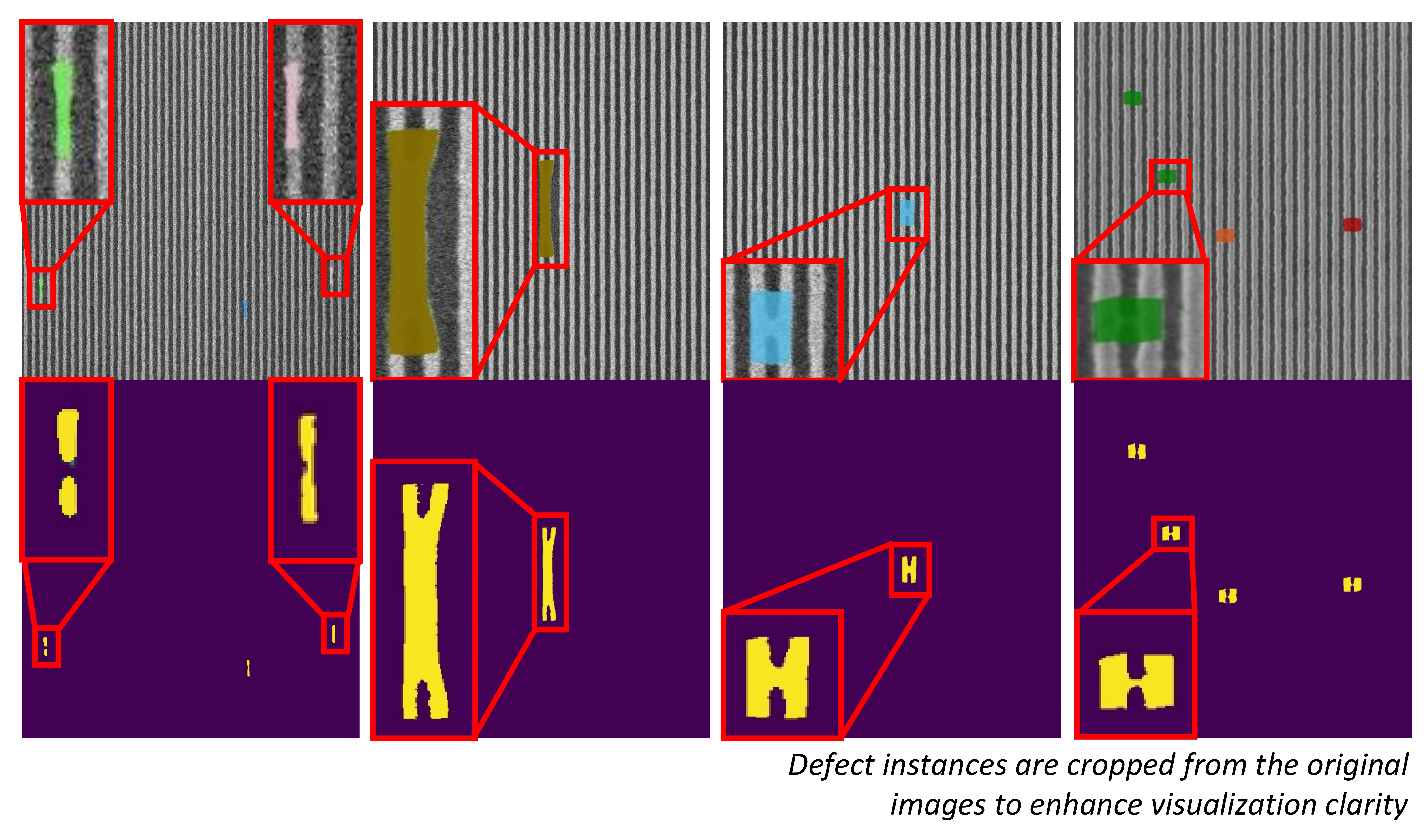}
        \caption{Left to right: gap and probable gap, line collapse, bridge, and micro bridge}
        \label{fig:results_adcds_adi}
    \end{subfigure}\\
    \begin{subfigure}{\textwidth}
        \centering
        \includegraphics[scale=0.14]{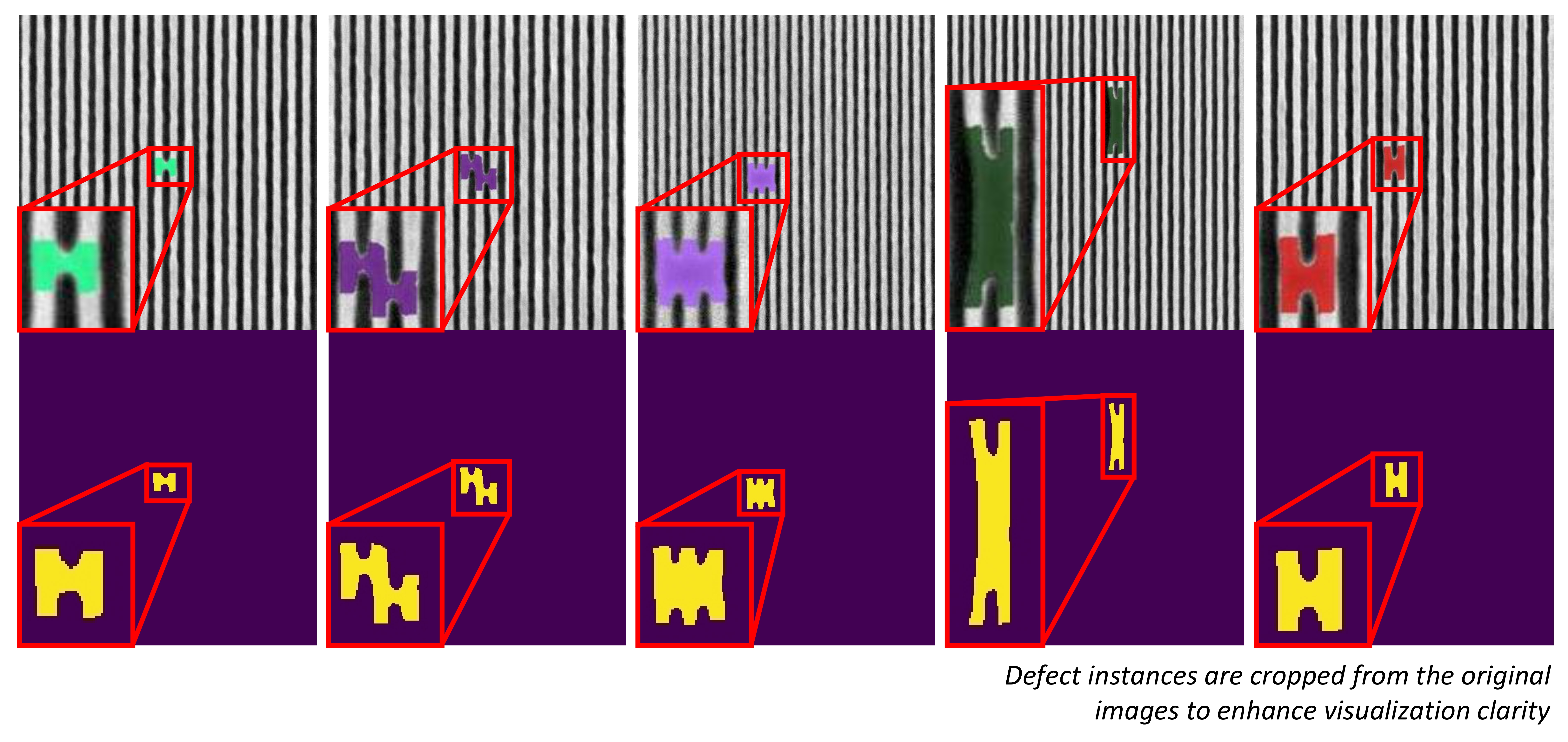}
        \caption{Left to right: thin bridge, MBNH, MBH, line collapse, single bridge}
    \end{subfigure}
    \caption{Defect mask segmentation and generation results with proposed ADCDS framework (BoxSnake based defect segmentation module) on ADI (a) and AEI (b) SEM datasets.}
    \label{fig:results_adcds}
\end{figure}

As we mentioned in previous section \ref{sec:training}, we explored each baseline configuration outlined in the original BoxSnake \cite{boxsnake} GitHub\footnote{\url{https://github.com/Yangr116/BoxSnake}} repository for defect segmentation module, and only depicted the best performing model configuration as Swin-Transformer Large.
The per-class AP and mAP metrics for this model configuration on both ADI and AEI SEM datasets are summarized in table \ref{tab:results_boxsnake}.
As previously mentioned, because ground-truth pixel-wise segmentation annotations for the ADI dataset were unavailable, we manually computed the necessary metrics to determine the ``segmentation per-class AP and mAP,'' focusing specifically on this dataset. These calculations were conducted solely based on the corresponding ground-truth bbox annotations available to us and are presented in table \ref{tab:tpfpboxsnake}. 
Manually computed segmentation AP and mAP for ADI SEM defect types/dataset is added to table \ref{tab:results_boxsnake} following the method as described in section \ref{sec:methodology}.
The inference timings (ms/image) vary and depend on the number of defects in a single SEM image, table \ref{tab:inferencespeed_box} shows the typical range for inference timings. 
However, it is understandable that ADI SEM images pose greater challenges compared to AEI SEM images due to factors such as low SNR, varying grayscale pixel contrast, and pixel-level extent, thus necessitating more time for segmentation.
Examples illustrating defect classification and segmentation results on both ADI and AEI SEM test datasets are presented in figure \ref{fig:results_adcds}. 
\begin{table}[htp]
    \caption{Per class AP and mean AP for Deformable DETR with a ResNet 50 backbone at different IoUs for ADI SEM dataset}
    \centering
    \fontsize{7.6pt}{9pt}
    \selectfont
    \begin{tabular}{|c|c||c|c|c|c|c||p{2.5cm}|}
        \hline
        \multirow{2}{*}{\textbf{\# queries}} & \multirow{2}{*}{\textbf{Epoch}} & \multicolumn{5}{c||}{\textbf{Per class AP @ IoU 0.5:0.95 (\%)}}  & \textbf{mAP @}\\ \cline{3-7}
        & & Microbridge & Bridge & Gap & Probable gap & Line collapse & \textbf{IoU 0.5:0.95 (\%)} \\
        \hline
        5 & 200 & 36.45 & 23.24 & 0.58 & 0.00 & 80.12 & \multicolumn{1}{c|}{28.08}\\
        \hline
        35 & 200 & 43.67 & 42.17 & 14.44 & 4.05 & 81.49 & \multicolumn{1}{c|}{ 37.16} \\
        \hline
        \cellcolor{gray!25}100 & \cellcolor{gray!25}200 & \cellcolor{gray!25}50.49 & \cellcolor{gray!25}42.44 & \cellcolor{gray!25}34.16 & \cellcolor{gray!25}13.53 & \cellcolor{gray!25}81.19 &\multicolumn{1}{c|}{ \cellcolor{gray!25}44.36} \\
        \hline
        & & \multicolumn{5}{c||}{\textbf{Per class AP @ IoU 0.5 (\%)}}  & \textbf{mAP @ IoU 0.5 (\%)} \\ \cline{3-7}
        \hline
        5 & 200 & 61.79 & 77.31 & 2.45 & 0.00 & 100.00 & \multicolumn{1}{c|}{48.31 } \\
        \hline
        35 & 200 & 72.72 & 77.23 & 36.08 & 12.10 & 100.00 & \multicolumn{1}{c|}{59.62} \\
        \hline
        \cellcolor{gray!25}100 & \cellcolor{gray!25}200 & \cellcolor{gray!25}76.63 & \cellcolor{gray!25}86.92 & \cellcolor{gray!25}68.84 & \cellcolor{gray!25}28.64 & \cellcolor{gray!25}99.93 & \multicolumn{1}{c|}{\cellcolor{gray!25}72.19} \\
        \hline
    \end{tabular}
    \label{tab:results_adi}
    \centering
    \fontsize{7.6pt}{9pt}
    \selectfont
    \caption{Per class AP and mean AP for Deformable DETR with a ResNet 50 backbone at different IoUs for AEI SEM dataset}
    \begin{tabular}{|c|c||c|c|c|c|c||p{2.5cm}|}
        \hline
        \multirow{2}{*}{\textbf{\# queries}} & \multirow{2}{*}{\textbf{Epoch}} & \multicolumn{5}{c||}{\textbf{Per class AP @ IoU 0.5:0.95 (\%)}}  & \textbf{mAP @}\\ \cline{3-7}
        & & Thin bridge & Single bridge & MBNH & MBH & Line collapse & \textbf{IoU 0.5:0.95 (\%)} \\
        \hline
        \cellcolor{gray!25}5 & \cellcolor{gray!25}200 & \cellcolor{gray!25}76.01 & \cellcolor{gray!25}62.34 & \cellcolor{gray!25}45.84 & \cellcolor{gray!25}35.80 & \cellcolor{gray!25}71.50 & \multicolumn{1}{c|}{\cellcolor{gray!25}58.30} \\
        \hline
        35 & 100 & 71.29 & 56.34 & 42.40 & 37.05 & 76.77 & \multicolumn{1}{c|}{56.77} \\
        \hline
        & & \multicolumn{5}{c||}{\textbf{Per class AP @ IoU 0.5 (\%)}}& \textbf{mAP @ IoU 0.5 (\%)} \\ \cline{3-7}
        \hline
        \cellcolor{gray!25}5 & \cellcolor{gray!25}200 & \cellcolor{gray!25}98.65 & \cellcolor{gray!25}92.93 & \cellcolor{gray!25}80.50 & \cellcolor{gray!25}79.84 & \cellcolor{gray!25}100.00 & \multicolumn{1}{c|}{\cellcolor{gray!25}90.38} \\
        \hline
        35 & 100 & 99.63 & 86.65 & 77.96 & 67.38 & 100.00 & \multicolumn{1}{c|}{86.33} \\
        \hline
    \end{tabular}
    \label{tab:results_aei}
    \centering
    \fontsize{7.6pt}{9pt}
    \selectfont
    \caption{Manually computed metrics to determine the segmentation per-class AP and mAP for BoxSnake with a Swin-Transformer Large backbone on ADI SEM defects}
    \begin{tabular}{|c|c|c|c|c|c|p{1.8cm}|}
   \hline
      & Microbridge & Bridge & Gap & Probable gap & Line collapse & Segmentation mAP @ IoU 0.5 (\%) \\ \hline
        Total instances & 47 & 19 & 156 & 49 & 66 & \\ \cline{1-6}
        False negative & 8 & 0 & 1 & 17 & 0 & \\ \cline{1-6}
        False positives & 9 & 0 & 9 & 17 & 0 & \\ \cline{1-6}
        True positives & 39 & 19 & 147 & 31 & 66 & \\ \hline
        Per class AP @ IoU 0.5 (\%) & 72.58 & 100.00 & 91.30 & 30.43 & 100.00 & \multicolumn{1}{|c|}{78.86}\\ \hline
    \end{tabular}
    \label{tab:tpfpboxsnake}
\end{table}
\begin{table}[htp]
    \caption{Detection inference speed for the ADI and AEI datasets}
    \centering
    \fontsize{7.6pt}{9pt}
    \selectfont
    \begin{tabular}{|c|c|c|}
        \hline
        \textbf{Dataset}    & \textbf{Model}  & \textbf{Inference speed (ms/image)}\\
        \hline
        \multirow{1}{*}{ADI} & D-DETR with 100 queries & 33.99 \\
        \hline
        \multirow{1}{*}{AEI} & D-DETR with 5 queries & 33.26 \\
        \hline
    \end{tabular}
    \label{tab:inferencespeed}
    \caption{Segmentation inference speed for the ADI and AEI datasets}
    \centering
    \fontsize{7.6pt}{9pt}
    \selectfont
    \begin{tabular}{|c|c|c|}
        \hline
        \textbf{Dataset}    & \textbf{Model}  & \textbf{Inference speed (ms/image)}\\
        \hline
        \multirow{1}{*}{ADI} & BoxSnake SwinL & 1500.00 - 10760.00 \\
        \hline
        \multirow{1}{*}{AEI} & BoxSnake SwinL & 180.00 - 320.00 \\
        \hline
    \end{tabular}
    \label{tab:inferencespeed_box}
\end{table}
\begin{table}[htp]
    \caption{Per class AP and mean AP for BoxSnake with a Swin-Transformer Large backbone on ADI and AEI SEM datasets}
    \centering
    \fontsize{7.6pt}{9pt}
    \selectfont
  \begin{tabular}{|c||c|c|c|c|c||p{2cm}|}
    \hline
    \multirow{3}{*}{\textbf{Dataset}} & \multicolumn{5}{c||}{\centering{\textbf{Per class AP @ IoU 0.5:0.95 (\%)}}}  & \textbf{mAP @} \\ \cline{2-6}
    & Thin bridge & Single bridge & MBNH & MBH & Line collapse & \textbf{IoU 0.5:0.95 (\%)} \\
    \hline
   & 68.90 & 70.62 & 34.03 & 38.77 & 71.68 &   \multicolumn{1}{|c|}{\textbf{56.80}} \\ \cline{2-7}
    \textbf{AEI}  & \multicolumn{5}{c||}{\textbf{Per class AP @ IoU 0.5 (\%)}} & \textbf{mAP @ IoU 0.5 (\%)} \\ \cline{2-7}
    & 99.53 & 96.40 & 84.25 & 97.21 & 100.00 &  \multicolumn{1}{|c|}{\textbf{95.48}}  \\
    \hline
   & \multicolumn{5}{c||}{\textbf{Per class AP (\%)}} & \textbf{mAP @ IoU 0.5 (\%)} \\ \cline{2-6}
    \textbf{ADI}  & Microbridge & Bridge & Gap & Probable gap & Line collapse &  \\
   \cline{2-7}
     & 72.58 & 100.00 & 91.30 & 30.43 & 100.00 &  \multicolumn{1}{|c|}{\textbf{78.86}} \\
    \hline
\end{tabular}
    \label{tab:results_boxsnake}
\end{table}

\section{Discussion} \label{sec:discussion}
\subsection{Proposed ADCDS framework}
In this section, we will delve into a comparative analysis of our proposed ADCDS framework, encompassing each corresponding module, in the realm of nano-scale defect inspection and segmentation. 
This comparison will be conducted against selected previous research works \cite{comparative_study,semi_centernet,semi_pointrend,semi_diffusioninst} that underwent benchmarking on the same real FAB dataset (for both ADI and AEI), supported by proof-of-validation. 
This will also aid us in identifying limitations of our proposed framework, thereby guiding further improvements and future research directions.
\vspace{-0.5cm}
\subsection{Comparative analysis of defect detection module against previous research works}
The reported results from SEMI-CenterNet \cite{semi_centernet} and the comparative study \cite{comparative_study} are summarised and compared against our proposed ADCDS framework in table \ref{tab:bbox_refs_res}, for both datasets. On the AEI SEM dataset, the Deformable DETR based proposed defect detection module and SEMI-CenterNet, both are trained for $200$ epochs. 
Our proposed defect detection module based on Deformable DETR demonstrates superior performance in all defect categories, achieving a mAP improvement of $51.24$\% @ IoU0.5.
For the ADI SEM dataset, it's evident that our model is underperforming, especially in smaller defect features such as gap and probable gap. However, Deformable DETR outperforms YOLOv7x in detecting instances of bridge by $20.62$\% and line collapse defects.\par
During our experimentation, we observed a decrease in AP for AEI defects (see table \ref{tab:results_aei}) when using more queries, whereas the opposite trend was noticed for ADI defects. 
This variation could be attributed to the number of defect instances present in the corresponding SEM datasets. 
The AEI dataset typically contains at most $1$ defect instance per image, and these defect features are generally more prominent and discernible to learn. 
Therefore, attempting to predict more instances in a single image may result in a decrease in AP. 
Conversely, the ADI dataset often exhibits a higher population density of defect instances (specifically, for gaps, probable gaps and micro-bridges) in a single image, tentatively up to $40$ defects. 
Additionally, learning features from ADI images is challenging due to factors such as noise interference, subtle grayscale variations between background and foreground etc. Therefore, increasing the number of queries improves both precision and recall.\par
SEMI-CenterNet \cite{semi_centernet} demonstrates a significant superiority over Deformable DETR in terms of inference speed metric, with $8.7$ ms/image compared to $33.26$ ms/image, for AEI SEM dataset. 
In contrast, for the ADI SEM dataset, Deformable DETR surpasses DINO \cite{comparative_study}, achieving $33.99$ ms/image compared to $108.7$ ms/image (with ResNet-50 backbone), and it is comparable to YOLOv7X with $20.3$ ms/image. 
However, YOLO architectures are single-stage object detector models and thus are expected to have faster inference speed.
\vspace{-0.5cm}
\subsection{Comparative analysis of defect instance segmentation module against previous research works}
SEMI-PointRend \cite{semi_pointrend} and SEMI-DiffusionInst \cite{semi_diffusioninst} all investigated segmentation applications on a similar AEI SEM dataset. The segmentation metrics, achieved by our proposed framework compared to these two previous benchmarks \cite{semi_pointrend,semi_diffusioninst} are presented in table \ref{tab:segm_results_ref}, expressed as per-class AP and mAP @ IoU 0.5:0.95. 
We also reported the same @ IoU 0.5. It is evident that proposed framework underperforms against previous frameworks on similar SEM dataset. However, it's essential to consider a conceptual validation here. The defect instance segmentation module based on BoxSnake \cite{boxsnake} is supervised solely by bounding box annotations, unlike the previous two architectures \cite{semi_pointrend,semi_diffusioninst}, which undergo supervised training with corresponding ground-truth segmentation annotations. 
We only utilized these ground-truth segmentation annotations to present a quantification metric, validating the performance of our proposed ADCDS framework in nano-scale defect instance segmentation solely guided by bounding box annotations provided by the previous D-DETR-based detection module.\par
To validate our rationale, we showcased the effectiveness of our proposed defect segmentation module on the ADI SEM dataset, where no precise ground-truth pixel-wise segmentation annotations exist. 
Figure \ref{fig:results_adcds_adi} visually illustrates the importance of our proposed framework, while table \ref{tab:tpfpboxsnake} presents the quantitative metrics. In table \ref{tab:tpfpboxsnake}, the AP is manually computed by considering the precision value at a specific recall level, typically set at 0.5, as AP = P@R=0.5. 
Finally, the mean Average Precision (mAP) is calculated by averaging the AP values across all classes or instances.\par
This is our first ever demonstration on nano-scale defect instance segmentation on this challenging ADI SEM dataset.
Segmenting gap and probable gap instances is particularly crucial because while bounding boxes can localize the occurrence of these critical defects, they do not accurately represent the pixel-wise defect extent. 
This pixel-wise characterization is essential for reducing stochastic defect rates and increasing device yield.
\begin{table}[h]
    \caption{Comparison analysis on defect detection (for both ADI and AEI SEM dataset) using proposed ADCDS framework against selected previous research works \cite{comparative_study,semi_centernet}.}
    \centering
    \fontsize{7.5pt}{9pt}
    \selectfont
    \begin{tabular}{|p{2cm}||c|c|c|c|c||c|p{0.7cm}|}
        \hline
        \multirow{2}{*}{\textbf{Model}}  & \multicolumn{5}{c||}{\textbf{ADI | Per class AP @ IoU 0.5 (\%)}}  & \textbf{mAP @ IoU} & \textbf{mAP @}\\ \cline{2-6}
         & Microbridge & Bridge & Gap & Probable gap & Line collapse & \textbf{0.5:0.95 (\%)} & \textbf{IoU 0.5 (\%)} \\
        \hline
        DINO \cite{comparative_study} & 82.4 & \textbf{96.0} & 96.0 & 57.9 & \textbf{100.00} & - & \textbf{86.5}\\
        \hline
        YOLOv7x \cite{comparative_study} & \textbf{85.4} & 66.3 & \textbf{97.4} & \textbf{72.1} & 99.5 & - & 83.5\\
        \hline
        Proposed ADCDS framework & 76.63 & 86.92 & 68.84 & 28.64 & 99.93 & \textbf{44.36} & 72.19 \\
        \hline
         & \multicolumn{5}{c||}{\textbf{AEI | Per class AP @ IoU 0.5:0.95 (\%)}}  & & \\ \cline{2-6}
         & Thin bridge & Single bridge & MBH & MBNH & Line collapse & & \\
        \hline
        SEMI-CN \cite{semi_centernet} & 6.95 & 32.60 & 20.17 & 0.00 & 57.75 & 23.49 & 39.14\\
        \hline
        Proposed ADCDS framework & \textbf{76.01} & \textbf{62.34} & \textbf{45.84} & \textbf{35.80} & \textbf{71.50} & \textbf{58.30} &  \multicolumn{1}{c|}{\textbf{90.38}}\\
        \hline
    \end{tabular}
    \label{tab:bbox_refs_res}
\end{table}
\vspace{-1.5cm} 
\begin{table}[h]
    \caption{Comparison analysis on defect segmentation (for AEI SEM dataset only) using proposed ADCDS framework against selected previous research works \cite{semi_pointrend,semi_diffusioninst}. Bold number represent best value.}
    \centering
    \fontsize{7.5pt}{9pt}
    \selectfont
    \begin{tabular}{|c||c|c|c|c|c||p{1.7cm}|}
        \hline
        \multirow{2}{*}{\textbf{Model}}    & \multicolumn{5}{c||}{\textbf{Per class AP @ IoU 0.5:0.95 (\%)}}  & \textbf{mAP @} \\ \cline{2-6}
        & Thin bridge & Single bridge & MBH & MBNH & Line collapse & \textbf{IoU 0.5:0.95 (\%)} \\
        \hline
        SEMI-PR \cite{semi_pointrend} & 55.0 & \textbf{77.7} & \textbf{53.5} & \textbf{57.4} & 63.6 & \multicolumn{1}{c|} {61.7}\\
        \hline
        SEMI-DI \cite{semi_diffusioninst} & \textbf{75.38} & 74.22 & 53.09 & 50.75 & \textbf{74.28} &\multicolumn{1}{c|}{\textbf{63.0}} \\
        \hline
        Proposed ADCDS framework & 68.90 & 70.62 & 34.03 & 38.77 & 71.68 & \multicolumn{1}{c|}{56.80}\\
        \hline
    \end{tabular}
    \label{tab:segm_results_ref}
\end{table} \\

\section{Conclusion}
\vspace{-0.2cm}
This research introduces a unified end-to-end framework for classifying, detecting, and segmenting multiple instances of semiconductor defects in aggressive pitches. Our framework integrates two modules. 
The detection module's role is to localize nano-scale defect instances in input SEM images across different process steps and provide that information, in terms of bounding boxes, to the segmentation module. 
The segmentation module's role is to generate pixel-precise segmented binary masks for various inter-class and intra-class defect instances to aid advanced semiconductor process control. 
Our proposed framework facilitates box-supervised defect instance segmentation, completely eliminating the laborious requirement for ground-truth pixel-wise mask annotation at the nano-scale level by human experts. 
We have benchmarked our proposed ADCDS framework on two real wafer SEM datasets for two different process steps, ADI and AEI, with a primary focus on the ADI SEM dataset. 
This marks the first demonstration of the applicability and significance of our proposed work on the ADI SEM dataset. 
This progress moves us towards unsupervised/weakly supervised defect instance segmentation, achieving 78.86\% mAP on ADI SEM images and 95.48\% on AEI SEM images with IoU@0.5. 
In the future, it can serve as a tool in the semiconductor industry for advanced defect analysis for advanced defect analysis in High Volume Manufacturing (HVM).

\bibliographystyle{splncs04}
\bibliography{references}
\end{document}